\documentclass{article}
\usepackage{ijcai19}

\usepackage{times}
\usepackage{soul}
\usepackage{url}
\usepackage[hidelinks]{hyperref}
\usepackage[utf8]{inputenc}
\usepackage[small]{caption}
\usepackage{graphicx}
\usepackage{amsmath}
\usepackage{booktabs}
\usepackage{algorithm}
\usepackage{algorithmic}

\usepackage{amsfonts}
\usepackage{subcaption}
\usepackage{enumitem}

\DeclareMathOperator*{\argmax}{argmax}
\DeclareMathOperator{\Cost}{Cost}

\title{Semi-Supervised Few-Shot Learning for Dual Question-Answer Extraction}

\author{
Jue Wang$^1$\and
Ke Chen$^1$\and
Lidan Shou$^1$\and
Sai Wu$^1$\And
Sharad Mehrotra$^2$\\
\affiliations
$^1$College of Computer Science and Technology, Zhejiang University\\
$^2$School of Information and Computer Science, University of California, Irvine\\
\emails
\{wangjuezju, chenk, should, wusai\}@zju.edu.cn, sharad@ics.uci.edu
}

\begin{document}

\maketitle

\begin{abstract}
This paper addresses the problem of key phrase extraction from sentences.
Existing state-of-the-art supervised methods require large amounts of annotated data to achieve good performance and generalization. 
Collecting labeled data is, however, often expensive.
In this paper, we redefine the problem as question-answer extraction, and present \textbf{SAMIE}: \textbf{S}elf-\textbf{A}sking \textbf{M}odel for \textbf{I}nformation \textbf{E}xtraction, a semi-supervised model which dually learns to ask and to answer questions by itself.
Briefly, given a sentence $s$ and an answer $a$, the model needs to choose the most appropriate question $\hat q$; meanwhile, for the given sentence $s$ and same question $\hat q$ selected in the previous step, the model will predict an answer $\hat a$. 
The model can support few-shot learning with very limited supervision. It can also be used to perform clustering analysis when no supervision is provided.
Experimental results show that the proposed method outperforms typical supervised methods especially when given little labeled data.
\end{abstract}

\section{Introduction}

Information Extraction (IE) refers to a spectrum of tasks which automatically extract structured information from unstructured texts. 
One important task in IE is to extract \emph{key phrases} and their respective \emph{categories} from sentences. Given a sentence $s$, the task needs to extract a ($c$,$p$)-pair, where $p$ is a phrase belonging to $s$ and $c$ indicates the category of $p$.
State-of-the-art approaches to this task employ supervised methods based on deep learning.
However, these approaches assume that there is sufficient annotated data, which in fact is not available in most situations.
Actually acquiring labels is costly, and probably the biggest obstacle to the application of these methods. This motivates the need for effective semi-supervised learning techniques leveraging unlabeled data.

\begin{figure}
\centering
\vspace*{0.2in}
\includegraphics[width=\columnwidth]{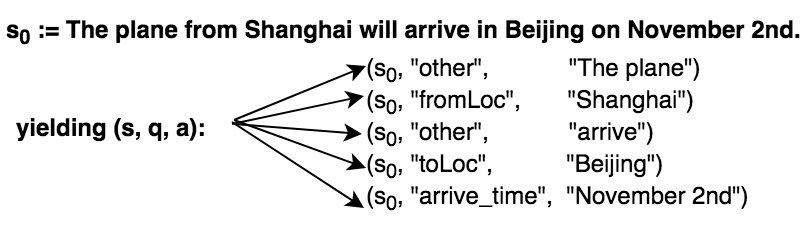}
\caption{(s, q, a)-triplets of a given sentence}
\label{fig:triple}
\end{figure}

By considering a question $q$ as a category, and an answer $a$ to it as a phrase, we define a new problem called \emph{dual question-answer extraction} to address key phrase extraction via question-answering. 
The new problem is described as: Given a sentence $s$ and an answer $a$, retrieve an appropriate question $\hat q$ for it; meanwhile, given sentence $s$ and question $q$, predict an answer $\hat a$ to $q$. The above sub-problems are named Question Selection (QS) and Answer Extraction (AE) respectively. In effect, our problem extracts ($q$,$a$)-pairs for a given sentence $s$, as illustrated in figure \ref{fig:triple}.
 

The main challenge comes from the lack of labeled \((s, q, a)\)-triplets. While a conventional solution requires a large number of labeled triplets, we advocate a novel approach where the need for labeled data can be significantly reduced.
In fact, a large number of \((s,a)\)-pairs without \(q\) can be easily collected, as it is reasonable to believe that almost all phrases in a sentence can be regarded as answers, i.e. given a sentence, there is always a question corresponding to each phrase (answer) in the sentence.
Since no supervision is needed in this step, \((s, a)\)-pairs can be regarded as unlabeled data.
The unlabeled $(s,a)$-pairs are useful in theory, because if there is a model to predict question $\hat q$ based on answer $a$, we can compensate the lack of \(q\) by acquiring \((s, \hat q, a)\) from \((s, a)\). 

\begin{figure*}
\centering
\vspace*{0.1in}
\includegraphics[width=0.7\textwidth]{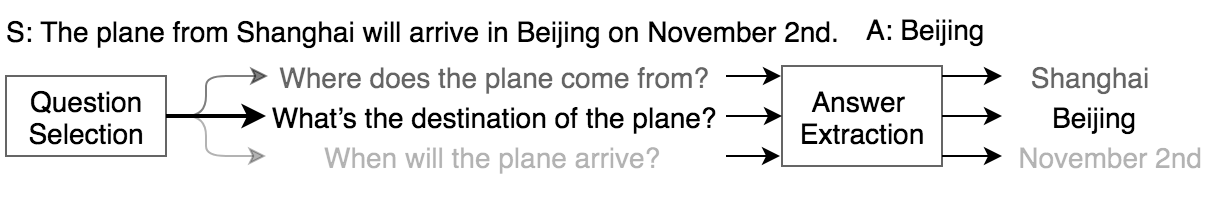}
\caption{Illustration of how \textit{SAMIE} works.}
\label{fig:example}
\end{figure*}

In this paper, we propose a novel neural network model called \textbf{SAMIE}: \textbf{S}elf-\textbf{A}sking \textbf{M}odel for \textbf{I}nformation \textbf{E}xtraction, which learns to ask and answer jointly with limited labeled data thus dually extracting question-answer pairs from given sentences.
The model includes two parts of training: the first part refers to a regular supervised method, where the QS sub-model and the AE sub-model are trained individually with labeled \((s, q, a)\)-triplets; in the second part, the QS sub-model predicts questions \(\hat q\) based on \((s, a)\)-pairs, and then the AE sub-model will give answers based on \((s, \hat q)\), so the supervision merely comes from \((s, a)\)-pairs. 
Specifically, by learning how to ask with numerous \((s, a)\), the model is forced to find patterns of answers in sentences, which reinforces its abstraction and generalization ability and prevents the model from overfitting on a small number of \((s, q, a)\)-triplets.

Figure \ref{fig:example} illustrates an example of how the model works.
Given a sentence as ``The plane from Shanghai will arrive in Beijing on November 2nd'', any phrase in it can be a potential answer. While the QS sub-model gives scores to all candidate questions (the higher the score, the darker the color in the figure), the AE sub-model gives answers to all candidate questions. But only the one whose question has the highest score is the relevant answer, which is ``Beijing''. 

It is possible to train without any \((s, q, a)\)-triplets (neglecting the supervised part of \textit{SAMIE}), but how the model converges depends on its initialization.
In fact, without pre-training or supervision, it may mistake the category and thus will be reduced to a clustering method.
To avoid this issue and enhance robustness, the model requires some supervision to ensure the direction it converges in.

We have evaluated our method on public datasets.
Experimental results show that the proposed method significantly outperforms traditional supervised methods especially in the case of extreme lack of labeled \((s, q, a)\)-triplets.

The key contributions of our work are as following:
\begin{itemize}
\item  We propose a framework that redefines key phrase extraction. Different from the restrictive problem definition in the literature which conducts key-phrase extraction relying on fixed categories, we study a more open problem of dual question-answer extraction.
\item We present a semi-supervised model for dual question-answer extraction. Supplied by very limited labeled sentences (below 500), the model can be effectively trained (F1$\approx$0.95) on large datasets of \((s,a)\)-pairs.
\item We demonstrate that the model can be used to perform phrase clustering without supervision.
\end{itemize}

\section{Proposed Method}

\begin{figure*}
\centering
\vspace*{0.2in}
\includegraphics[width=1.0\textwidth]{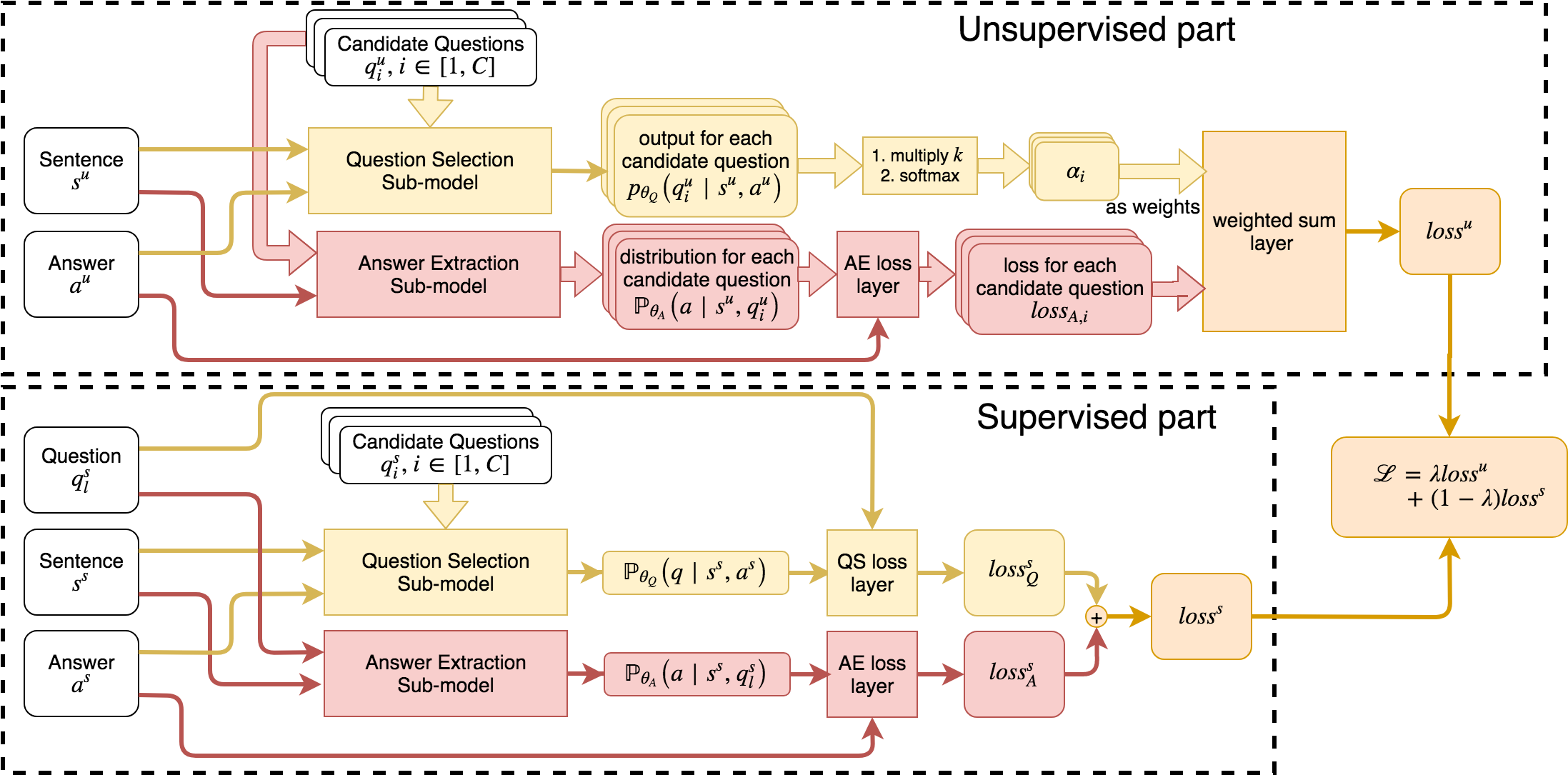}
\caption{The \textit{SAMIE} framework. The yellow part is related to question selection and the red part is related to answer extraction. Unsupervised part and supervised parts are outlined by dashed lines in the figure.}
\label{fig:framework}
\vspace*{0.1in}
\end{figure*}

\subsection{Problem definition}

Note \(\mathcal S,\mathcal Q, \mathcal A\) as spaces of sentences,
questions and answers respectively and \(s \in \mathcal S\),
\(q \in \mathcal Q\), \(a \in \mathcal A\).

The dual question-answer extraction task can be represented as a triplet \((s, q, a)\), consisting of a sentence \(s\), a question \(q\) and an answer \(a\) to \(q\). Note that the answer is a phrase or a single word in the sentence, which requires the ability to exploit context information in both sentence and question. 

There are two sub-problems to be solved, namely \emph{Question Selection} and \emph{Answer Extraction}.

\subsubsection{Question Selection}

The model needs to evaluate the relevance of each candidate question to the given sentence and answer. The most appropriate question for the given sentence and answer should be the most relevant one.

More formally, given a sentence \(s\) and an answer \(a\), our model's main objective is to learn the conditional probability distribution \(p_{\theta_Q}(q|s,a)\) where \(\theta_Q\) indicates all parameters related to QS, to select the most appropriate question \(\hat q\) from the candidate questions provided by us:
\begin{equation}
\hat q = \argmax_{q} p_{\theta_Q}(q | s, a)
\end{equation}

\subsubsection{Answer Extraction}

For the AE task, the model aims at predicting an answer to the given sentence and question. 
Because we assume the answer is included in the given sentence, the task can be reduced to a conditional sequence labeling problem, where the condition is given by a question.

Formally, given a sentence \(s\) and a question \(q\), the goal is to learn the conditional probability distribution \(p_{\theta_A}(a|s,q)\) where \(\theta_A\) indicates all parameters related to AE, to extract the correct answer \(\hat a\):
\begin{equation}
\hat a = \argmax_{a} p_{\theta_A}(a | s, q)
\end{equation}

\subsection{Dataset Preparation}

When preparing the data, it is necessary to note that, although \((s,a)\)-pairs are not directly available, collecting them is not expensive. With raw sentences available, the collection of \((s, a)\)-pairs can be obtained using existing POS taggers and named entity recognizers e.g. Stanford CoreNLP\footnote{https://stanfordnlp.github.io/CoreNLP/}.
The collection allows for some missing possible answers, as the performance will not be affected much. But it cannot include too many errors on the boundaries of phrases.

Without pre-training or using well-trained word embeddings, the model may not converge to the position as we wish, i.e. the definitions of questions may be mistaken. 
An approach to overcome this challenge is to provide a small sample of \((s, q, a)\)-triplets as supervision.

We need to define and give \emph{groups} for questions that are written in natural language. Each group contains questions expressing the same intent. Although each group is allowed to contain as few as only one question, it is recommended to include more for robust understanding of questions. 
During the training phase, for each query, one question is picked randomly from each group forming the candidate questions.

\subsection{Framework}
Figure \ref{fig:framework} presents the framework of \textit{SAMIE}.
The QS sub-model predicts the probability of a question given a sentence and an answer, noted \(p_{\theta_Q}(q|s, a)\) (or \(\mathbb P_{\theta_Q}(q|s, a)\) representing its distribution).
Meanwhile, the AE sub-model predicts the distribution of answers given a sentence and a question, noted \(\mathbb P_{\theta_A}(a|s, q)\) (or \(p_{\theta_A}(a|s, q)\) indicating the probability of a specific answer).
Note that \(\theta_Q \cap \theta_A \not\equiv \emptyset\), i.e. they can share parameters such as the embeddings of $s$, $q$, and $a$.

During the training phase, the framework has two parts: the supervised part trained on \((s,q,a)\) where variables are marked with ``s'' and the unsupervised part trained on \((s,a)\) where variables are marked with ``u''.

There are \(C\) categories of candidate questions, as mentioned, questions in the same category express the same intent but are in different forms. The candidate questions mainly serve for QS sub-model which will give possibilities of the candidate questions and find the most appropriate question for the given sentence and answer.

Particularly, in the unsupervised part, candidate questions are also fed into AE sub-model yielding \(C\) answers, where only one of them matches the input answer \(a^u\). With the help of QS sub-model which produces probabilities for all candidate questions, the answers are weighted during loss computing with the irrelevant ones mitigated. As a result, the loss can be used to guide the iterative learning of both the QS and AE sub-models. 
In other words, considering each sub-model to define a ``hyper-dimension'', we treat the parameter spaces of QS and AE as orthogonal hyper-dimensions, so they can both be trained by the loss simultaneously.

\subsection{Training Method}

\subsubsection{Supervised part}

The supervised part computes the sum of the losses of two sub-tasks, with the following notations
\begin{itemize}
  \item \(s^s \in \mathcal S\): the input sentence
  \item \(a^s \in \mathcal A\): the input answer
  \item \(\{q^s_1, q^s_2, ..., q^s_C\} \in \mathcal Q^C\): the candidate questions
  \item \(q^s_l\): the labeled question
  where \(l \in [1, C]\) i.e. \(q^s_l\) is among \(\{q^s_i\}_{i \in [1,C]}\)
\end{itemize}

Then \(\mathbb P_{\theta_Q}(q|s^s, a^s)\) and \(\mathbb P_{\theta_A}(a|s^s, q^s)\) can be obtained from the QS sub-model and the AE sub-model, where
\begin{equation}
\begin{gathered}
  \mathbb P_{\theta_Q}(q|s^s, a^s) = \sum_{i=1}^C p_{\theta_Q}(q|s^s, a^s) \mathbb{I}_{\{q^s_i\}}(q) \\
  \mathbb{I}_{X}(x) =
  \left\{
    \begin{gathered}
      1, \quad \text{if } x \in X \\
      0, \quad \text{if } x \not \in X
    \end{gathered}
  \right.
\end{gathered}
\end{equation}

And the total loss is given as
\begin{equation}
\begin{aligned}
loss^s &= loss^s_Q + loss^s_A \\
&= \Cost_Q(q^s_l, \mathbb P_{\theta_Q}(q|s^s, a^s)) \\
&+ \Cost_A(a^s, \mathbb P_{\theta_A}(a|s^s,q^s_l))
\end{aligned}
\end{equation}

\subsubsection{Unsupervised part}

In the unsupervised part, we do not have any labeled question, and the prediction of QS sub-model is actually an intermediate state in the training phase.
With notations as following
\begin{itemize}
  \item \(s^u \in \mathcal S\): the input sentence
  \item \(a^u \in \mathcal A\): the input answer
  \item \(\{q^u_1, q^u_2, ..., q^u_C\} \in \mathcal Q^C\): the candidate questions
\end{itemize}
the loss of the unsupervised part is computed as
\begin{align}
loss^u = \sum_{i=1}^{C} {\alpha_i \Cost_A(a^u, \mathbb P_{\theta_A}(a|s^u, q^u_i))}, \\
\alpha_i = \frac{\exp(k p(q^u_i | s^u, a^u))}{\sum^C_{j=1}{\exp(k p(q^u_j | s^u, a^u))}}
\end{align}
The softmax layer is required to avoid getting a trivial solution i.e.~\(p_{\theta_Q}(q|s,a) \equiv 0\) thus
\(loss \equiv 0\).
And the output of QS sub-model should be multiplied by $k\ge1$ before the softmax layer, making the entropy of output of the softmax layer smaller, emphasizing the possible question and ignoring the irrelevant ones. Otherwise, all questions may be given too much attention, causing that QS sub-model hesitates and updates continuously while AE sub-model cannot learn anything. Eventually, the model may not converge.

\subsubsection{Joint training}

During the training phase, we train the models jointly by minimizing the total loss: 
\begin{equation}
    \mathcal L = \lambda loss^u + (1 - \lambda) loss^s
\end{equation}
where \(\lambda > 0\) is a parameter giving the importance of the unsupervised part. When \(\lambda\) is close to 0, the optimizer will focus more on the supervised part; when \(\lambda\) is close to 1, the optimizer will focus more on the unsupervised part. 
Because variables are initialized randomly, the unsupervised part may be very confused at the beginning. So it turns out better to begin with a small \(\lambda\), i.e.~pay more attention to the supervised part, and then increase \(\lambda\) gradually as training time goes on.

\subsection{Evaluation}

\subsubsection{Question Selection}

Given a sentence and its answer \((s, a)\), as well as all candidate
questions \(\{q_i\}_{i \in [1, C]}\), the model will predict which
question is the most appropriate one. The selected question is:
\begin{equation}
\begin{aligned}
\hat {q} &= \argmax_q p_{\theta_Q}(q | s, a) \\
&= q_{\argmax_{i=1,2,...,C} p_{\theta_Q}(q_i | s, a)}.
\end{aligned}
\end{equation}

When \(s\) and \(a\) are irrelevant i.e \(a\) is not a valid answer of
\(s\), no question should be selected. Thus a threshold
\(p_{\text{th}}\) should be set, where \begin{equation}
p_{\theta_Q}(\hat q|s,a) > p_{\text{th}}.
\end{equation}
Otherwise, no question would be selected.

\subsubsection{Answer Extraction}

Given a sentence and a question, the relevant answer is predicted as
\begin{equation}
\hat a = \argmax_a p(a|s,q)
\end{equation}

When \(s\) and \(q\) are irrelevant, no answer should be extracted. This should be taken into account because a sentence may not contain all kind of slots i.e. answers cannot always be found in the sentence for all questions. However, the AE sub-model will always try its best to find the most possible answer no matter which kind of question is posed i.e. it always produces an answer.

Therefore we should check the result by verifying
\begin{equation}
p(q|s,\hat a) > p_{\text{th}}.
\end{equation}
Otherwise, no answer would be extracted.

\subsection{Implementation}

The input layer is able to represent a single text sentence where each word will be mapped to a high-dimensional vector space to obtain a fixed embedding.
For a given word, its input representation is constructed by summing the corresponding word and position embeddings.
Positional embeddings \cite{gehring2017convolutional} are required as we use encoder and decoder of the Transformer \cite{vaswani2017attention}, which contains no recurrence and convolution.

We denote the multi-layer bidirectional Transformer encoder and decoder as \(\text{Encoder}\) and \(\text{Decoder}\).
All \(\text{Encoder}\)s share the same word and position embeddings in our implementation, and the \(\text{Decoder}\) is used to merge two token sequences into one. 

The sentence, question and answer should first be encoded to capture the contextual semantic information.
\begin{equation}
s^* = \text{Encoder}_s(s),
q^* = \text{Encoder}_q(q),
a^* = \text{Encoder}_a(a)
\end{equation}

In our setting, the answer is a short phrase in a sentence, so the AE task is a sequence labeling task. The AE sub-model can be then implemented with \(\text{Decoder}\) followed by a fully-connected layer: 
\begin{align}
h_A &= \text{Decoder}_0(s^*, q^*) \\
logits_A &= \text{Dense}(h_A)
\end{align}

The logits represent the non-normalized possibility of whether a word of the original sentence should be regarded as a part of the answer, and all words whose logit is positive form the predicted answer.

We expect the QS sub-model to give the relevance of \(q\) and \((s, a)\). Therefore, \(q\) and \((s, a)\) are encoded to two vectors and then computing their cosine similarities. We use the following procedure to get the relevance of \(q\) and \((s, a)\):
\begin{align}
v_Q &= \text{mean}(q^*) \\
v_{S,A} &= \text{mean}(\text{Decoder}_1(s^*, a^*)) \\
score &= \text{cosine}(v_Q, v_{S,A})
\end{align}
$\Cost_Q$ and $\Cost_A$  can be arbitrary losses defined for multiclass classification and over two sequences. Both of them can use the prevalent cross-entropy.


\section{Experiments}

\begin{figure*}
\centering
\vspace*{0.1in}
  \begin{subfigure}[t]{0.5\columnwidth}
    \centering
    \includegraphics[width=\textwidth]{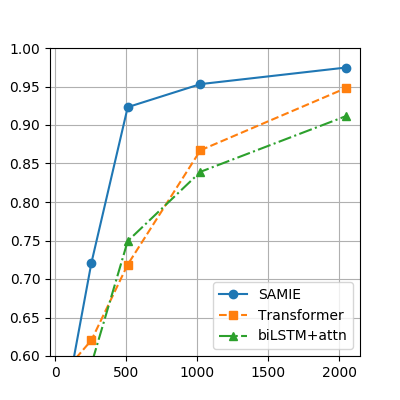}
    \caption{ACC - Question Selection}
  \end{subfigure}
  \begin{subfigure}[t]{0.5\columnwidth}
    \centering
    \includegraphics[width=\textwidth]{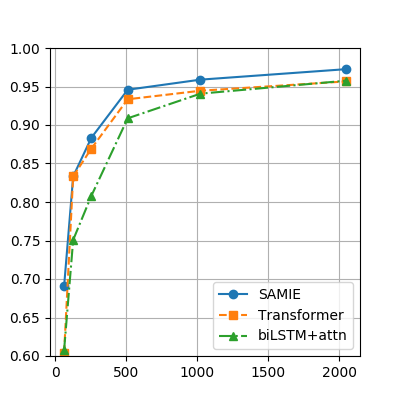}
    \caption{Precision - Answer Extraction}
  \end{subfigure}
  \begin{subfigure}[t]{0.5\columnwidth}
    \centering
    \includegraphics[width=\textwidth]{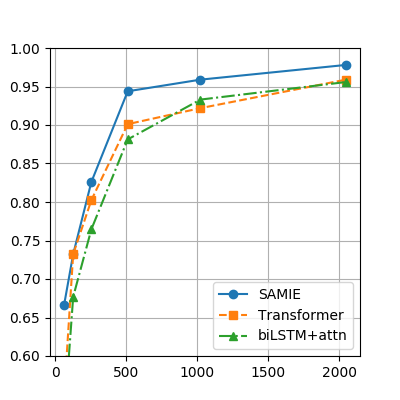}
    \caption{Recall - Answer Extraction}
  \end{subfigure}
  \begin{subfigure}[t]{0.5\columnwidth}
    \centering
    \includegraphics[width=\textwidth]{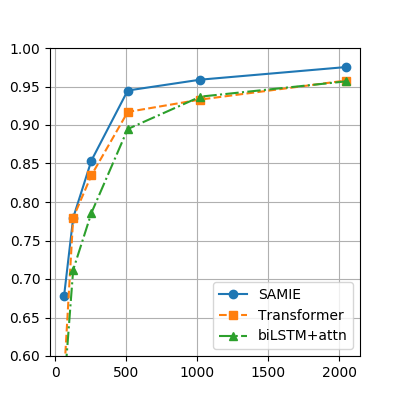}
    \caption{F1 - Answer Extraction}
  \end{subfigure}
\caption{Metrics along with the number of annotated sentences on ATIS.}
\label{fig:experiments}
\end{figure*}

\subsection{Dataset}

We evaluate on the Airline Travel Information System (ATIS) dataset and Chinese Emergency Corpus (CEC). 
The former contains spoken queries on flight related information, associated intents and key slots in the statements; the latter contains news about emergencies with key elements and their properties. 
We extract slots of each query by extracting question-answer pairs.
To compare the differences between different methods more fairly and clearly, we have filtered out data that does not meet the requirements, and the remaining data still accounts for the majority of the original data.
As these datasets do not originally contain any question-answer pair, a group of questions are given by ourselves for each type of slot. 
The question types for ATIS includes ``airline'', ``arrive\_time'', ``depart\_time'', ``return\_time'', ``fromloc'', ``toloc'', and ``stoploc''; and those for CEC are ``time'', ``location'', ``denoter'', ``participant''.

\subsection{Comparative Models}

We compare our approach to the following classical supervised learning methods.
\begin{itemize}
  \item biLSTM+attn: For the QS task, use bidirectional LSTM and attention layer to encode the given sentence and answer and predict the most relevant question; for the AE task, use bidirectional LSTM to encode the given sentence and question, and use attention to combine the two sequences before output layer.
  \item Transformer: Similar to the above one where biLSTM is replaced with Transformer.
  \item SAMIE: Our proposed model (an implementation of the framework with Transformer), trained on \((s, q, a)\)-triplets and \((s, a)\)-pairs.
  \item $*$ (Small): Any model marked with ``Small'' is a small version of the original one with fewer units, fewer layers therefore fewer trainable parameters.
\end{itemize}
Note that the Transformer-based approach has the same QS sub-model and AE sub-model as \textit{SAMIE}.

\subsection{Evaluation Settings}

Only a small amount of data is picked as labeled \((s, q, a)\)-triplets. Note that one sentence may yield several \((s, q, a)\)-triplets, thus we can extract over ten thousand triplets from ATIS and about eight thousand triplets from CEC. We randomly select 15\% data from the two datasets as their respective testsets.

We test all models with 64, 128, 256, 512, 1024, 2048 labeled sentences which are converted to \((s, q, a)\)-triplets for further training needs. 
The additional \((s, a)\)-pairs for \textit{SAMIE} are all prepared from the remaining trainset.
In addition, we conduct a detailed study on the 512 labeled sentences to evaluate \textit{SAMIE} against overfitting.

We evaluate the QS task and the AE task separately. Given sentences with answers, the QS sub-model should predict their corresponding questions; and given sentences with questions, the AE sub-model should extract possible answers for them. Note that the precision, recall, and F1-score for answer extraction are \emph{word-based} rather than slot-based. So they should be a little higher than the slot-based ones.

In the end, we train \textit{SAMIE} without labeled \((s, q, a)\)-triplets, and check the confusion matrix to see the effect of our method in phrase clustering.

\subsection{Results}

\begin{table}
\centering
\begin{tabular}{ccc}
\\ 
\toprule
   & QS & AE \\
   & \begin{tabular}{c}Acc\end{tabular}
   & \begin{tabular}{ccc}prec & recall & F1\end{tabular}
   \\
\midrule
biLSTM+attn (Small)
& \begin{tabular}{c}0.73\end{tabular}
& \begin{tabular}{ccc}0.91 & 0.92 & 0.92\end{tabular}      \\
biLSTM+attn (Regular)
& \begin{tabular}{c}0.75\end{tabular}
& \begin{tabular}{ccc}0.91 & 0.88 & 0.90\end{tabular}      \\
Transformer (Small)
& \begin{tabular}{c}0.71\end{tabular}
& \begin{tabular}{ccc}0.93 & 0.91 & 0.92\end{tabular}      \\
Transformer (Regular)
& \begin{tabular}{c}0.72\end{tabular}
& \begin{tabular}{ccc}0.93 & 0.90 & 0.91\end{tabular}      \\
\midrule
SAMIE (Small)
& \begin{tabular}{c}0.81\end{tabular}
& \begin{tabular}{ccc}0.88 & 0.88 & 0.88\end{tabular}      \\
SAMIE (Regular)
& \begin{tabular}{c}0.92\end{tabular}
& \begin{tabular}{ccc}0.95 & 0.94 & 0.95\end{tabular}      \\
\bottomrule
\end{tabular}
\caption{Detailed comparison with 512 labeled sentences on ATIS.}
\label{tab:experiments}
\end{table}

Due to space limit, we only present the results on the ATIS dataset, but our approach also has similar performance on CEC (the accuracy reaches 0.9 for the QS task with about 500 triplets).
Figure \ref{fig:experiments} shows performances of different methods on ATIS. 
Among them, the QS task needs special attention, because when using \((s, a)\)-pairs, the choice of question is only an intermediate state without any supervision and if questions were selected incorrectly, the error would be propagated to AE sub-model. Therefore, the improvement of \textit{SAMIE} in the QS task illustrates that it has learned well. 
On the other hand, the result of AE also indicates the contribution of leveraging \((s,a)\)-pairs to the model.

Due to the limited training resources, all baselines in supervised learning did not perform well. However, \textit{SAMIE}, leveraging \((s, a)\)-pairs, had significant improvements compared to others. When the amount of data is small, the gap between them is particularly noticeable. The results confirm \textit{SAMIE} as a promising solution to quickly extract information of interest at very limited expense in human annotation.
As the number of labeled triplets increases, the performance margins between \textit {SAMIE} and the baseline methods tend to narrow down. However, it is still slightly better than the rest, even when the number of labeled ones becomes very close to the total number (3439) of sentences.

As the amount of labeled data is very small, the problem of overfitting is inevitable. However, our approach has good performance in generalization.
This is verified in the detailed performance study of 512 labeled sentences, as shown in table \ref{tab:experiments}. 
It is known that conventional models (such as Transformer) typically address the problem of overfitting by reducing the number of variables of the network. This method is confirmed by the results of Transformer (Small) and Transformer (Regular) in the AE task, as the former outperforms the latter. However, this method weakens the network and restricts the final performance.
Nevertheless, \textit{SAMIE} alleviates the problem of overfitting. 
Specifically, \textit{SAMIE} (Regular) outperforms \textit{SAMIE} (Small) on all metrics, while both still beat their respective Transformer counterparts. These results reveal that \textit{SAMIE} works better with more powerful neural network.

\begin{figure}
  \centering
  \includegraphics[width=1.0\columnwidth]{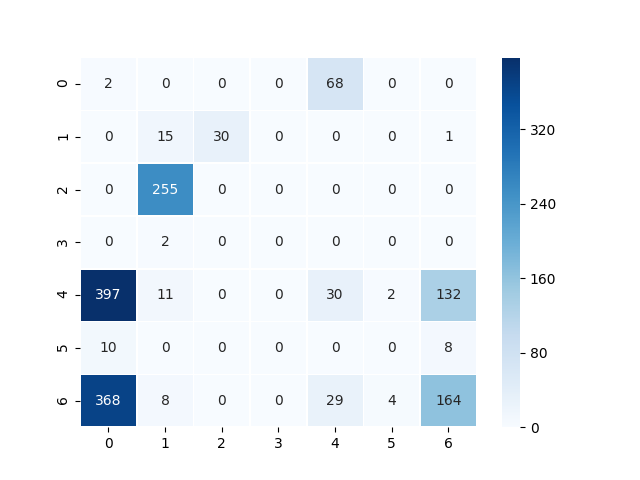}
  \caption{Confusion matrix without labeled triplets on ATIS}
  \label{fig:confusion}
\end{figure}

Finally, we show the possibility to perform a clustering analysis with our model. 
We trained \textit{SAMIE} without any labeled \((s, q, a)\)-triplets i.e. only the unsupervised part was used, and the confusion matrix shown in figure \ref{fig:confusion} demonstrates that, although the model made mistakes in the results, it only placed \((s, a)\)-pairs into limited wrong clusters.
The model mainly confused ``fromloc'' and ``stoploc'', a mistake understandable due to the similarity between the two categories. 


\section{Related Work}

The IE task is usually treated as a sequence labeling problem in which contiguous sequences of words are assigned semantic class labels. 
Standard approaches to solving this problem include generative models, such as maximum entropy Markov models (MEMMs) \cite{mccallum2000maximum} and conditional random fields (CRFs) \cite{raymond2007generative}.
Deep learning approaches such as Recurrent Neural Networks (RNNs) have attracted much attention because of their superior performance in language modeling and understanding tasks. Many researchers \cite{mesnil2013investigation,mesnil2015using,yao2013recurrent,yao2014spoken} applied RNNs to this task and have promising achievements. \cite{liu2016attention} added attention mechanism \cite{bahdanau2014neural} on top of RNNs and further improved the accuracy.
However, although RNN-based approaches have been proved effective in this kind of problem, they are relatively slow due to their failure in parallelization.
Similar work has been done in CNNs \cite{xu2013convolutional} which are parallel processable.

These approaches have had significant achievements, but they had a strong assumption that there is enough labeled data.
Without enough labeled data, these models failed to perform well and the problem of over-fitting becomes especially severe.
Furthermore, the categories of key phrases (or slots) should be predefined and are difficult to expand.


Some researchers cast IE problem to reading comprehension. \cite{levy2017zero} reduced relation extraction to answering simple reading comprehension questions. Specifically, given a sentence with an annotated \((h, r, t)\)-triplet where \(h,t\) are entities and \(r\) is their relation, it models the \(h,r\) as a question, and the left \(t\) as its answer. \cite{roth2018neural} further improved the model performance on a similar problem setting. Furthermore, based on previous works, \cite{qiu2018qa4ie} focused on cross-sentence relation argument extractions.

There are some studies which explore the relationship between asking and answering to enhance each other.
\cite{golub2017two,kumar2018automating} used the two-stage process generating questions based on answers another network selects, formally, they factorized \(\mathbb P(q, a|p)\) into \(\mathbb P(a|p) \mathbb P(q|p, a)\) which were computed by two networks. \cite{tang2017question} used regularizations to connect the two networks and trained them simultaneously, which improved performances of both. \cite{wang2017joint} used a single seq2seq model to generate questions and answers, whose idea is simple but very effective. \cite{sachan2018self} applied self-learning strategy for jointly learning to ask as well as answer questions, leveraging unlabeled text along with labeled question-answer pairs for learning.

\section{Conclusion}

In this paper, we studied the task of extracting question-answer pairs.
The problem was addressed by solving two sub-tasks: Question Selection and Answer Extraction.
We observed that these two sub-tasks are intrinsically linked and that \((s,a)\)-pairs are much easier to acquire than \((s, q, a)\)-triplets. 
Therefore, we proposed \textit{SAMIE}, a semi-supervised learning model that can be trained by \((s, a)\)-pairs and a small number of labeled \((s, q, a)\)-triplets. 
Our experimental results showed that \textit{SAMIE} worked especially well when given very small amount of labeled data, and that it outperformed the baseline methods.
In the future, we would consider to further reduce the need for labeled data and try to achieve unsupervised learning with the help of pre-training techniques. 


\newpage

\bibliographystyle{named}
\bibliography{paper}

\begin{thebibliography}{}

\bibitem[\protect\citeauthoryear{Bahdanau \bgroup \em et al.\egroup
  }{2014}]{bahdanau2014neural}
Dzmitry Bahdanau, Kyunghyun Cho, and Yoshua Bengio.
\newblock Neural machine translation by jointly learning to align and
  translate.
\newblock {\em CoRR}, abs/1409.0473, 2014.

\bibitem[\protect\citeauthoryear{Gehring \bgroup \em et al.\egroup
  }{2017}]{gehring2017convolutional}
Jonas Gehring, Michael Auli, David Grangier, Denis Yarats, and Yann~N Dauphin.
\newblock Convolutional sequence to sequence learning.
\newblock {\em arXiv preprint arXiv:1705.03122}, 2017.

\bibitem[\protect\citeauthoryear{Golub \bgroup \em et al.\egroup
  }{2017}]{golub2017two}
David Golub, Po-Sen Huang, Xiaodong He, and Li~Deng.
\newblock Two-stage synthesis networks for transfer learning in machine
  comprehension.
\newblock {\em arXiv preprint arXiv:1706.09789}, 2017.

\bibitem[\protect\citeauthoryear{Kumar \bgroup \em et al.\egroup
  }{2018}]{kumar2018automating}
Vishwajeet Kumar, Kireeti Boorla, Yogesh Meena, Ganesh Ramakrishnan, and
  Yuan-Fang Li.
\newblock Automating reading comprehension by generating question and answer
  pairs.
\newblock In {\em Pacific-Asia Conference on Knowledge Discovery and Data
  Mining}, pages 335--348. Springer, 2018.

\bibitem[\protect\citeauthoryear{Levy \bgroup \em et al.\egroup
  }{2017}]{levy2017zero}
Omer Levy, Minjoon Seo, Eunsol Choi, and Luke Zettlemoyer.
\newblock Zero-shot relation extraction via reading comprehension.
\newblock In {\em Proceedings of the 21st Conference on Computational Natural
  Language Learning (CoNLL 2017)}, pages 333--342, 2017.

\bibitem[\protect\citeauthoryear{Liu and Lane}{2016}]{liu2016attention}
Bing Liu and Ian Lane.
\newblock Attention-based recurrent neural network models for joint intent
  detection and slot filling.
\newblock {\em arXiv preprint arXiv:1609.01454}, 2016.

\bibitem[\protect\citeauthoryear{McCallum \bgroup \em et al.\egroup
  }{2000}]{mccallum2000maximum}
Andrew McCallum, Dayne Freitag, and Fernando~CN Pereira.
\newblock Maximum entropy markov models for information extraction and
  segmentation.
\newblock In {\em Icml}, volume~17, pages 591--598, 2000.

\bibitem[\protect\citeauthoryear{Mesnil \bgroup \em et al.\egroup
  }{2013}]{mesnil2013investigation}
Gr{\'e}goire Mesnil, Xiaodong He, Li~Deng, and Yoshua Bengio.
\newblock Investigation of recurrent-neural-network architectures and learning
  methods for spoken language understanding.
\newblock In {\em Interspeech}, pages 3771--3775, 2013.

\bibitem[\protect\citeauthoryear{Mesnil \bgroup \em et al.\egroup
  }{2015}]{mesnil2015using}
Gr{\'e}goire Mesnil, Yann Dauphin, Kaisheng Yao, Yoshua Bengio, Li~Deng, Dilek
  Hakkani-Tur, Xiaodong He, Larry Heck, Gokhan Tur, Dong Yu, et~al.
\newblock Using recurrent neural networks for slot filling in spoken language
  understanding.
\newblock {\em IEEE/ACM Transactions on Audio, Speech, and Language
  Processing}, 23(3):530--539, 2015.

\bibitem[\protect\citeauthoryear{Qiu \bgroup \em et al.\egroup
  }{2018}]{qiu2018qa4ie}
Lin Qiu, Hao Zhou, Yanru Qu, Weinan Zhang, Suoheng Li, Shu Rong, Dongyu Ru,
  Lihua Qian, Kewei Tu, and Yong Yu.
\newblock Qa4ie: A question answering based framework for information
  extraction.
\newblock {\em arXiv preprint arXiv:1804.03396}, 2018.

\bibitem[\protect\citeauthoryear{Raymond and
  Riccardi}{2007}]{raymond2007generative}
Christian Raymond and Giuseppe Riccardi.
\newblock Generative and discriminative algorithms for spoken language
  understanding.
\newblock In {\em Eighth Annual Conference of the International Speech
  Communication Association}, 2007.

\bibitem[\protect\citeauthoryear{Roth \bgroup \em et al.\egroup
  }{2018}]{roth2018neural}
Benjamin Roth, Costanza Conforti, Nina Poerner, Sanjeev Karn, and Hinrich
  Sch{\"u}tze.
\newblock Neural architectures for open-type relation argument extraction.
\newblock {\em arXiv preprint arXiv:1803.01707}, 2018.

\bibitem[\protect\citeauthoryear{Sachan and Xing}{2018}]{sachan2018self}
Mrinmaya Sachan and Eric Xing.
\newblock Self-training for jointly learning to ask and answer questions.
\newblock In {\em Proceedings of the 2018 Conference of the North American
  Chapter of the Association for Computational Linguistics: Human Language
  Technologies, Volume 1 (Long Papers)}, volume~1, pages 629--640, 2018.

\bibitem[\protect\citeauthoryear{Tang \bgroup \em et al.\egroup
  }{2017}]{tang2017question}
Duyu Tang, Nan Duan, Tao Qin, Zhao Yan, and Ming Zhou.
\newblock Question answering and question generation as dual tasks.
\newblock {\em arXiv preprint arXiv:1706.02027}, 2017.

\bibitem[\protect\citeauthoryear{Vaswani \bgroup \em et al.\egroup
  }{2017}]{vaswani2017attention}
Ashish Vaswani, Noam Shazeer, Niki Parmar, Jakob Uszkoreit, Llion Jones,
  Aidan~N Gomez, {\L}ukasz Kaiser, and Illia Polosukhin.
\newblock Attention is all you need.
\newblock In {\em Advances in Neural Information Processing Systems}, pages
  5998--6008, 2017.

\bibitem[\protect\citeauthoryear{Wang \bgroup \em et al.\egroup
  }{2017}]{wang2017joint}
Tong Wang, Xingdi Yuan, and Adam Trischler.
\newblock A joint model for question answering and question generation.
\newblock {\em arXiv preprint arXiv:1706.01450}, 2017.

\bibitem[\protect\citeauthoryear{Xu and Sarikaya}{2013}]{xu2013convolutional}
Puyang Xu and Ruhi Sarikaya.
\newblock Convolutional neural network based triangular crf for joint intent
  detection and slot filling.
\newblock In {\em Automatic Speech Recognition and Understanding (ASRU), 2013
  IEEE Workshop on}, pages 78--83. IEEE, 2013.

\bibitem[\protect\citeauthoryear{Yao \bgroup \em et al.\egroup
  }{2013}]{yao2013recurrent}
Kaisheng Yao, Geoffrey Zweig, Mei-Yuh Hwang, Yangyang Shi, and Dong Yu.
\newblock Recurrent neural networks for language understanding.
\newblock In {\em Interspeech}, pages 2524--2528, 2013.

\bibitem[\protect\citeauthoryear{Yao \bgroup \em et al.\egroup
  }{2014}]{yao2014spoken}
Kaisheng Yao, Baolin Peng, Yu~Zhang, Dong Yu, Geoffrey Zweig, and Yangyang Shi.
\newblock Spoken language understanding using long short-term memory neural
  networks.
\newblock In {\em Spoken Language Technology Workshop (SLT), 2014 IEEE}, pages
  189--194. IEEE, 2014.

\end{thebibliography}

\end{document}